\documentclass{article}
\usepackage{ijcai24}

\usepackage{times}
\usepackage{soul}
\usepackage{url}
\usepackage[hidelinks]{hyperref}
\usepackage[utf8]{inputenc}
\usepackage[small]{caption}
\usepackage[pdftex]{graphics}
\usepackage{amsmath}
\usepackage{amsthm}
\usepackage{booktabs}
\usepackage{xcolor}
\usepackage{amsfonts,array,amssymb,epsfig,bm,diagbox,multirow,epstopdf,subfigure}
\usepackage[switch]{lineno}

\usepackage[ruled,linesnumbered]{algorithm2e}
\SetKwComment{Comment}{ $\triangleright$ }{}

\title{How Multimodal Integration Boost the Performance of LLM for Optimization: Case Study on Capacitated Vehicle Routing Problems}

\author{
Yuxiao Huang$^1$
\and
Wenjie Zhang$^1$\and
Liang Feng$^2$\and
Xingyu Wu$^1$\and
Kay Chen Tan$^1$\\
\affiliations
$^1$The Hong Kong Polytechnic University\\
$^2$Chongqing University\\
\emails
\{yuxiao.huang, wenjie-ee.zhang, xingy.wu, kctan\}@polyu.edu.hk,
liangf@cqu.edu.cn
}
\begin{document}

\maketitle

\begin{abstract}

Recently, large language models (LLMs) have notably positioned them as capable tools for addressing complex optimization challenges. Despite this recognition, a predominant limitation of existing LLM-based optimization methods is their struggle to capture the relationships among decision variables  when relying exclusively on numerical text prompts, especially in high-dimensional problems. Keeping this in mind, we first propose to enhance the optimization performance using multimodal LLM capable of processing both textual and visual prompts for deeper insights of the processed optimization problem. This integration allows for a more comprehensive understanding of optimization problems, akin to human cognitive processes. We have developed a multimodal LLM-based optimization framework that simulates human problem-solving workflows, thereby offering a more nuanced and effective analysis. The efficacy of this method is evaluated through extensive empirical studies focused on a well-known combinatorial optimization problem, i.e., capacitated vehicle routing problem. The results are compared against those obtained from the LLM-based optimization algorithms that rely solely on textual prompts, demonstrating the significant advantages of our multimodal approach.

  % Very recently, large language models (LLMs) have emerged as powerful instruments for addressing intricate optimization challenges. However, a salient limitation of existing LLM-based optimization techniques is their incapacity to grasp the interdependencies among decision variables of optimization problems when relying solely on numerical text prompts, particularly in high-dimensional settings. To address this shortcoming, we first propose to enhance the optimization performance by employing multimodal LLM that can handle both textual and visual prompts for profound comprehension of the optimization problem. Such integration enables a more holistic understanding of optimization problems, akin to human cognitive mechanisms. We have devised a multimodal LLM-based optimization scheme that emulates human problem-solving procedures, thus providing a more refined and effective analysis. The efficacy of this method is validated through rigorous empirical investigations centered on a well-known combinatorial optimization problem, namely, capacitated vehicle routing problem (CVRP). The results are contrasted with those attained from the LLM-based optimization algorithms that utilize only textual prompts, manifesting the considerable benefits of our multimodal approach.
\end{abstract}

\section{Introduction}
Optimization problems arise in various domains, such as supply chain management \cite{cox1999power}, logistics \cite{pfohl2010logistics}, engineering \cite{deb2012optimization}, and machine learning \cite{sra2012optimization}. These problems are typically solved by traditional heuristic search methods, such as evolutionary algorithm \cite{yu2010introduction}, ant colony algorithm \cite{blum2005ant}, simulated annealing \cite{bertsimas1993simulated}, and so on. Despite the success of these algorithms on various optimization problems, the no free lunch theory \cite{wolpert1997no} indicates that their operators are not universally effective, which means that they may perform poorly on newly arrived problems. Moreover, developing new algorithms for novel problems entails a rigorous and laborious process, which often requires much expert knowledge. Hence, it would be advantageous to have a more intelligent optimization engine, which can automatically acquire the features of the optimization problem and adopt suitable search strategies to enhance the search performance.

Very recently, large language models (LLMs) have emerged as a powerful technique for natural language understanding and generation \cite{bubeck2023sparks}, as well as for other modalities such as images and videos \cite{bubeck2023sparks,huang2023chatgpt}. By processing massive amounts of data and learning from diverse sources of information, LLMs are capable of performing various tasks across domains. Additionally, LLMs can be conveniently accessed and queried by using natural language prompts, which renders them attractive for users who lack familiarity with formal languages or programming. Inspired by these benefits, some efforts have been made to investigate the use of LLMs as optimizers, where the optimization problem is expressed in natural language and the LLM produces optimization solutions according to the prompts. For instance, Yang \textit{et al.} introduced Optimization by PROmpting (OPRO) that exploits LLMs to solve optimization problems \cite{yang2023large}. Li \textit{et al.} developed OptiGuide that uses LLMs to provide insights and guidance for supply chain optimization \cite{li2023large}. Liu \textit{et al.} proposed a novel framework that combines LLMs and evolutionary algorithms to solve multi-objective optimization problems\cite{liu2023large}. Moreover, Cummins \textit{et al.} used LLMs to generate compiler optimization sequences \cite{cummins2023large}. Pluhacek \textit{et al.} leveraged LLMs to generate novel meta-heuristic optimization algorithms that can adapt to different problem domains \cite{pluhacek2023leveraging}. Romera \textit{et al.} showed that LLMs can discover new mathematical theorems and proofs from natural language descriptions of mathematical problems \cite{romera2023mathematical}. These works demonstrate the remarkable ability of LLMs to learn from natural language and generate optimization solutions across different domains.

However, most existing LLM-based optimization methods adopt numerical text prompts to model the optimization problem, which restricts the LLMs' ability to capture the relationships between the decision variables. As the number of decision variables grows, the text prompts become longer and more complicated, which further impairs the problem representation and diminishes the performance of the LLMs. Nevertheless, the optimization problems in real-world frequently involve multiple forms of representation. For example, capacitated vehicle routing problem (CVRP) can be represented by text to show the information of each customer node, or by image to visualize the layout of the nodes.

Keeping the above in mind, we propose to use multimodal LLM (MLLM) for optimization, which can handle both textual and visual inputs for enhanced optimization performance. MLLMs leverage the complementary information from various modalities and offer more rich and expressive representations of the optimization problem at hand. In this work, we present a MLLM-based optimization framework, where the optimization problem is represented by a mix of textual and visual prompts, and the LLM generates optimization solutions based on both modalities. To evaluate the performance of the proposed method, the framework is applied to a well-known combinatorial optimization problem, i.e., CVRP, which entails assigning a set of customers to a fleet of vehicles with limited capacity, while minimizing the total traveling distance. Furthermore, We compare our method with the LLM-based optimization algorithm that uses only text input, and demonstrate that our method outperforms it in terms of solution quality. The main contributions of this paper are outlined as follows:
\begin{itemize}
\item The proposed method incorporates both textual and visual prompts simultaneously to facilitate the comprehensive understanding of optimization problems and further improve optimization performance. To the best of our knowledge, this is the first multi-modal solution for optimization.
\item A novel XML text prompt is proposed in this work for the optimization problem, which allows the MLLM model to capture more precise relationships among the decision variables.
\item With the assistance of LLM, the framework of MLLM has been specially designed to emulate the workflow of human beings in solving optimization problems. This endows MLLM with enhanced generalization capabilities, enabling its application to various optimization problems that can benefit from the introduction of graphical illustrations.
\item Using a well-known combinatorial optimization problem, CVRP, we evaluate our proposed framework that exploits both textual and visual inputs, and compare it with the ones that use only textual inputs.
\end{itemize}

The rest of the paper is organized as follows. The CVRP is introduced in Section 2 as a case study to assess the optimization efficacy of MLLM. The application of LLM to optimization problems is also surveyed in this section Furthermore, section 3 describes the proposed optimization framework based on MLLM. Subsequently, section 4 reports and analyzes the experimental outcomes of our method. Lastly, section 5 summarizes the paper and outlines the directions for future work.

\section{Background}
\subsection{Capacitated Vehicle Routing Problem}
\begin{figure}[htbp]
  \centering
  \includegraphics[width=0.8\columnwidth]{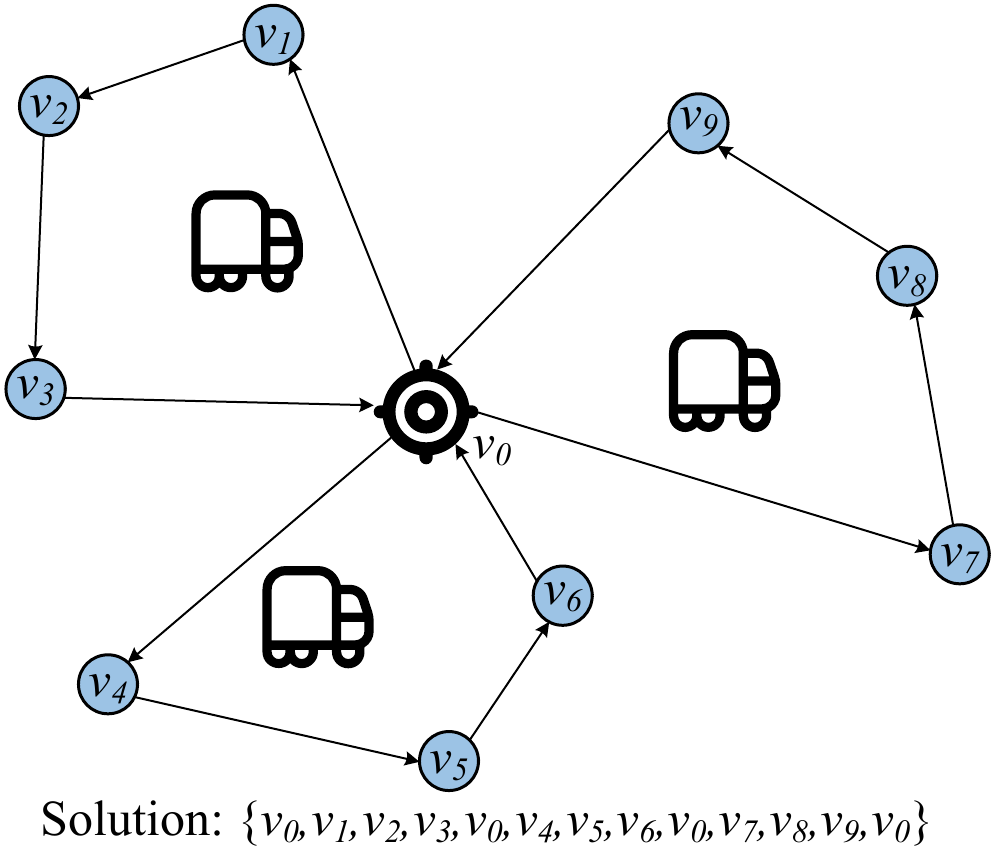}
  \caption{Illustration of a CVRP with optimal visiting routes.}
  \label{fig:CVRP}
\end{figure}

CVRP is a well-known combinatorial optimization problem that entails finding the optimal routes for a fleet of vehicles to serve a given set of customers (i.e., nodes) subject to the constraint of vehicle capacity \cite{toth2014vehicle}. The CVRP and its variants have numerous applications in various domains, such as smart logistics \cite{sajid2021novel} and distributing medical supplies for emergencies \cite{zheng2023multi}. Mathematically, let $\mathbf{G}=(\mathbf{E},\mathbf{V})$ be an undirected connection graph for a CVRP, where $\mathbf{E}$ and $\mathbf{V}$ denote the sets of edges and customers, respectively, and $\mathbf{D}$ is the demand set of customers. Given $K$ vehicles of capacity $Q$, the objective of the CVRP is to find a set of routes $\mathbf{R_k}, k=1,...,K$, each originating and terminating at a depot $v_0$, where the vehicles are stationed, such that:
\begin{itemize}
\item Every customer $v_{i} \in \mathbf{V}$ is served by exactly one vehicle and served only once.
\item Total demand of the customers on each route does not exceed the vehicle capacity: $\forall k, \sum_{v_i\in \mathbf{R}_k}\mathbf{D}(v_i)\le Q$.
\item The sum of the distances of all the routes is minimized.
\end{itemize}
As shown in Fig. \ref{fig:CVRP}, an optimal solution for a CVRP instance depends largely on the spatial distribution of customers. This motivates us to introduce the MLLM to handle both textual and visual inputs, and thus gain a deeper understanding of the problems require to solve.

\subsection{Literature Review}
Despite the LLMs have been applied to various problems in human beings' daily life, the study on using LLMs for optimization remains nascent. Existing LLM-assisted optimization methods can be briefly divided into two classes. The first class is to use LLMs as black-box optimizers that can generate new solutions from a numerical textual prompt, which contains the current solutions and their fitness values. For instance, OPRO \cite{yang2023large} employed LLMs to solve optimization problems without gradients. OptiGuide \cite{li2023large} leveraged LLMs to offer insights and guidance for supply chain optimization. Moreover, Meyerson \textit{et al.} \cite{meyerson2023language} devised Language Model Crossover, a simple technique to evolve semantically-rich text representations for various domains. Brownlee \textit{et al.} \cite{brownlee2023enhancing} assessed the use of LLMs as mutation operators to enhance the search process. However, since these methods use numerical prompts to guide the LLMs, they have difficulty in handling the complex relationships between decision variables, which thus hinders the LLMs' optimization efficacy on broader scopes with larger scales.

On the other hand, the second class is to use LLMs as meta-heuristics that can design novel functions or heuristics for solving complex problems, by leveraging their natural language processing abilities. In particular, Guo \textit{et al.} \cite{guo2023towards} used LLMs to optimize the hyperparameters of gradient-based methods, such as learning rate, momentum, and weight decay. Cummins \textit{et al.} \cite{cummins2023large} introduced LLMs to generate compiler optimization sequences that outperform human experts and state-of-the-art methods. Pluhacek \textit{et al.} \cite{pluhacek2023leveraging} leveraged LLMs to generate novel meta-heuristic optimization algorithms that can adapt to different problem domains. Moreover, Funsearch\cite{romera2023mathematical} employed LLMs to discover new mathematical functions that surpass human performance. However, the LLMs employed in these studies are solely prompted with textual data, which might hinder their ability to grasp the intricacies and nuances of the problem domain.

Bearing the above in mind, we first present a new framework that uses MLLM to help solve optimization problems, which can take advantage of both visual and textual information.

\section{Multimodal LLM for Optimization}

\begin{figure*}[htbp]
  \centering
  \includegraphics[width=2\columnwidth]{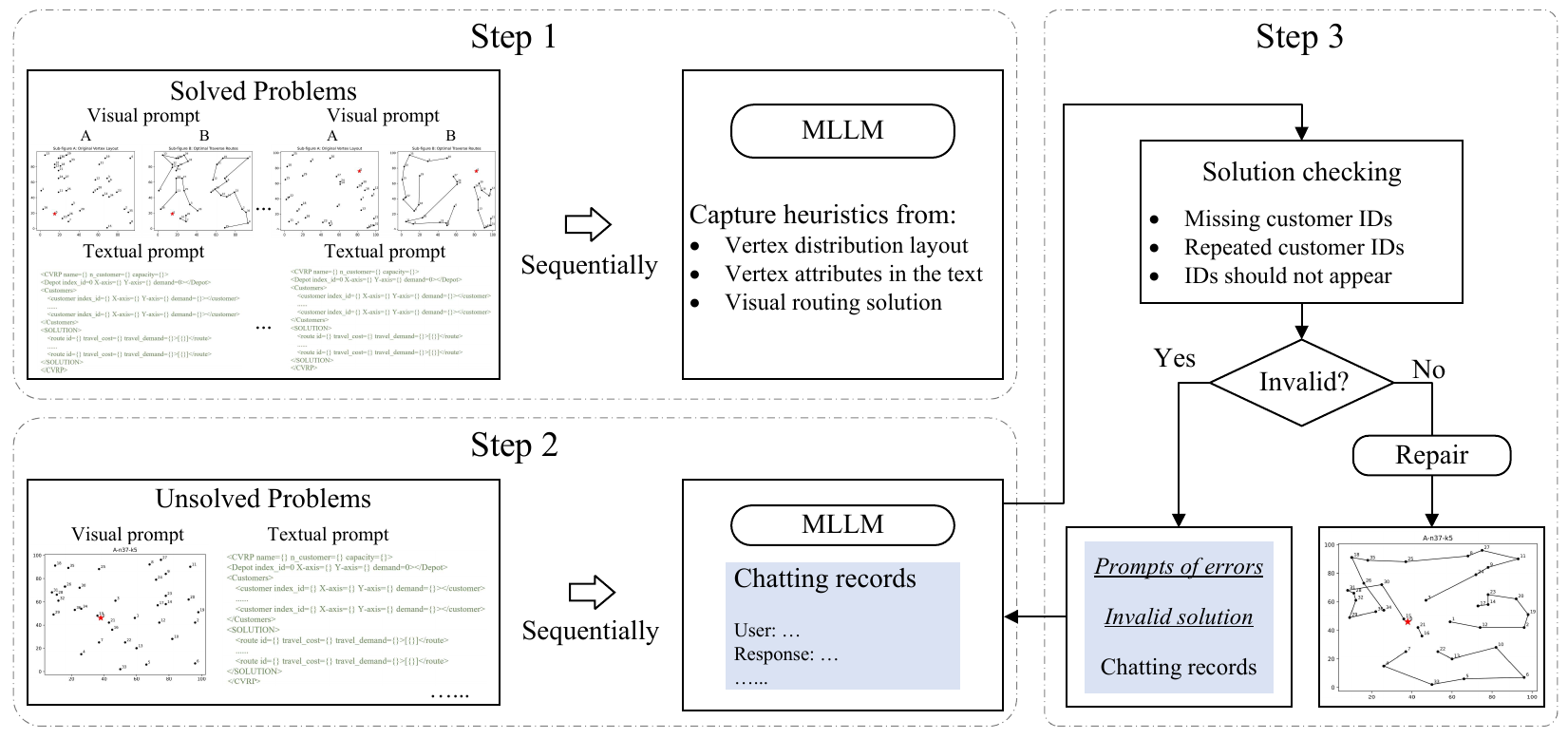}
  \caption{Workflow of our proposed method with three steps.}
  \label{fig:workflow}
\end{figure*}

In this section, we present our optimization framework based on MLLMs, which is a general approach that mirrors the nuanced steps inherent in human problem-solving methodologies applied to complex optimization scenarios. The entire workflow has been depicted in Fig .\ref{fig:workflow}. As can be observed, the framework encompasses three pivotal steps: 1) heuristic extraction from solved problems; 2) solution generation via learned heuristics; and 3) solution evaluation and refinement. In what follows, the details for each step are provided.

\subsection{Heuristic Extraction from Solved Problems}
The first step is to find the heuristics from solved similar problems for enhanced optimization on new problems. To this end, we propose to utilize the MLLM to capture deeper problem features with both textual and visual modalities, thereby facilitating more profound discernment of heuristics \cite{lenat1982nature} to effectively and efficiently solve the problems that are not yet observed.

As shown in Fig. \ref{fig:workflow}, we first provide MLLM some solved CVRPs that are similar to the unseen problem to handle. Each solved problem is accompanied by a well-crafted textual description and a visual representation of the vertex distribution layout, which consists of two sub-figures. The left part (marked as ``A'') shows the original layout, while the right part (marked as ``B'') illustrates the layout with the optimal traveling routes. This graphical modality enables MLLM to intuitively grasp the underlying heuristics for handling CVRPs. Next, the proposed method entails that the MLLM delivers its observations based on the textual and visual prompts, which are preserved as past chats for devising solutions to unsolved CVRPs.

In this study, we facilitate the MLLM's understanding of the problem by using the standard XML format to describe each solved CVRP, which is given by:
\begin{center}
\fcolorbox{black}{cyan!20}{\parbox{0.9\linewidth}{
You will help me create initial high-quality solutions for the Capacitated Vehicle Routing Problems (CVRPs)$\dots$\\
To create initial high-quality solutions for the new CVRPs, I will first show you some solved CVRPs as examples$\dots$\\
The following format is used to describe each solved CVRP with text information, where \{\} denotes variables:\\
$<$CVRP name=\{\} n\_customer=\{\} capacity=\{\}$>$ \\
$<$Depot$>$\dots$<$/Depot$>$ \\
$<$Customers$>$\dots$<$/Customers$>$ \\
$<$SOLUTION$>$\dots$<$/SOLUTION$>$ \\
$<$/CVRP$>$ \\
Now you will be provided with several solved CVRPs with the description of XML text and the figure with original topological layout and optimal traveling routes$\dots$\\
You may start by finding the accurate customer mapping between the XML document and the sub-figures according to the IDs, and then return the observations you found\dots
}}
\end{center}

As can be observed, the XML format provides a concise and clear summary of the problem properties of CVRPs. Furthermore, Fig. \ref{fig:workflow} shows that the sub-figures have vertex ID numbers annotated, which facilitates the MLLM to map the vertexes from textual data to visual data.

\subsection{Solution Generation via learned heuristics}
The second step of our proposed framework is to generate a preliminary solution via MLLM by applying the heuristics learned from the previous observations. To this end, we use MLLM again, but with a different input and output format. Particularly, we prompt the MLLM with the description of the problem to solve, which consists of the textual and visual prompts, without the optimal solution (the ``Solution'' in the XML text and the sub-figure ``B'' are removed). The MLLM is then expected to generate an initial solution in the form of an XML document, which contains only the ``Solution'' tag with the traveling routes. The heuristics assist the MLLM in generating the initial solution by providing it with guidance from the historical chats. However, the MLLM is not restricted by these historically derived heuristics, and it can also investigate new possibilities and produce novel solutions by applying its own knowledge. An example of this step has been provided in what follows. As can be observed, the prompts indicate that the MLLM should produce a brief and accurate routing solution that covers all the customers.

\begin{center}
\fcolorbox{black}{cyan!20}{\parbox{0.9\linewidth}{
Now you are presented with the unsolved CVRPs with the description of XML text (without $<$SOLUTION$>$\{\}$<$/SOLUTION$>$) and topological layout picture (without the figure of optimal traveling routes)\dots \\
$<$CVRP name=\{\} n\_customer=\{\} capacity=\{\}$>$ \\
$<$Depot$>$\dots$<$/Depot$>$ \\
$<$Customers$>$\dots$<$/Customers$>$ \\
$<$/CVRP$>$ \\
Kindly return me the complete preliminary solution of \dots in XML format, adhering to the heuristics that you have previously acquired. \\
No Explanation Needed. \\
}}
\end{center}

\subsection{Solution Evaluation and Refinement}
The third final step of the proposed framework is to evaluate and revise the solution produced by the MLLM based on the MLLM's response. As a generative model that relies on probabilistic inference, MLLM cannot guarantee the correctness and completeness of the routing solution it produces, particularly when the problem dimensionality increases. Hence, we iteratively guide the MLLM to correct the invalid result by supplying the errors detected in the solution. More specifically, after obtaining the generated solution from the MLLM using its learned heuristics, the solution will be passed to a validation function that verifies the customer IDs in the solution. The validation component will detect for the following errors: missing customer IDs, repeated customer IDs, and customer IDs that should not be in the solution. It then returns the invalid solution to the MLLM with the error prompts to assist the MLLM in correcting the solution. Once the MLLM produces a solution with valid customer IDs for each route, we use the ``Repair’’ operator to adjust the routes in case of capacity violation. This operator may create additional routes for serving the extra customers. We provide an example of this step in the following:

\begin{center}
\fcolorbox{black}{cyan!20}{\parbox{0.9\linewidth}{
Your routing solution is invalid. To return valid routes, refine the ones below by removing duplicate customer IDs and adding missing ones:\\
$<$SOLUTION$>$ \\
$<$route id=1\dots$>$[1,3,7,9,9]$<$/route$>$ \\
$<$route id=2\dots$>$[2,4,6,8,10,11]$<$/route$>$ \\
\dots \\
$<$/SOLUTION$>$ \\
The duplicated customer IDs are given by: [9], the missed customer IDs are given by: [5], and the customer IDs which should not appear are given by: [11] \\
Please remove the duplicated IDs and the IDs should not appear, and add the missed IDs to the route with minimum customers \\
No Explanations Needed \\
}}
\end{center}

Moreover, to facilitate the complete grasp of this study, an exhaustive illustration of chats is attached to the document as a reference.

\section{Experimental Study}
In this study, we propose a novel framework for solving the CVRP, which is a challenging combinatorial optimization problem with many real-world applications. To test the effectiveness of our proposed method, we use a set of widely adopted CVRP benchmarks\footnote{https://vrp.atd-lab.inf.puc-rio.br/index.php/en/} that vary in customer size and vertex layout. Specifically, we classify these benchmarks into two categories: small-scale problems with up to 50 nodes and large-scale problems with around 160 nodes. Additionally, we leverage three CVRPs with known optimal solutions, namely ``A-n45-k6'', ``P-n60-k10'' and ``E-n101-k14'', as knowledge prompts to facilitate the learning and reasoning process of the proposed method.

The problem properties of each CVRP instance have been tabulated in Table \ref{tab:benchmark}, where ``$|V|$'', ``$|K|$'' and ``Capacity'' denote the number of vertexes (depot involved), the number of vehicles and the capacity of each vehicle, respectively. To demonstrate the superiority of our proposed framework (namely MLLM-V), comparative analysis has been conducted against the baseline that solely relies on textual prompts (namely MLLM-T). It should be noted that, both methods employ a same multimodal language model, namely \textit{GPT-4-vision-preview}, without any further fine-tuning. The only difference between MLLM-V and MLLM-T lies in the use of visual prompts and their corresponding descriptions.

\begin{table}[htbp]
  \centering
  \caption{Problem properties of the employed CVRP instances.}
  \setlength{\tabcolsep}{6pt} % Default value: 6pt
  \renewcommand{\arraystretch}{0.9} % Default value: 1
  \begin{tabular}{llrrr}
  \toprule
  Benchmark & Problem & $|V|$ & $|K|$ & Capacity\\
  \midrule
  \multirow{7}*{Small} 
  & P-n19-k2 & 18 & 2 & 160 \\
  & A-n32-k5 & 32 & 5 & 100 \\
  & A-n36-k5 & 36 & 5 & 100 \\
  & A-n38-k5 & 38 & 5 & 100 \\
  & A-n39-k5 & 39 & 5 & 100 \\
  & A-n44-k6 & 44 & 6 & 100 \\
  & A-n46-k7 & 46 & 7 & 100 \\
  \midrule
  \multirow{10}*{Large}
  & E-n51-k5 & 51 & 5 & 160 \\
  & A-n65-k9 & 65 & 9 & 100 \\
  & A-n69-k9 & 69 & 9 & 100 \\
  & P-n55-k10 & 55 & 10 & 115 \\
  & P-n65-k10 & 65 & 10 & 130 \\
  & P-n70-k10 & 70 & 10 & 135 \\
  & X-n139-k10 & 139 & 10 & 106 \\
  & X-n143-k7 & 143 & 7 & 1190 \\
  & X-n153-k22 & 153 & 23 & 144 \\
  & X-n162-k11 & 162 & 11 & 1174 \\
  \midrule
  \multirow{3}*{Solved}
  & A-n45-k6 & 45 & 6 & 100 \\
  & P-n60-k10 & 60 & 10 & 120 \\
  & E-n101-k14 & 101 & 14 & 112 \\
  \bottomrule
  \end{tabular}
  \label{tab:benchmark}
\end{table}

\subsection{Results}
In this section, we present the results of our numerical experiments, where the performance of our proposed MLLM-V method is compared with the MLLM-T method on a set of representative CVRP benchmarks. Table \ref{tab:cvrp} summarizes the results of both methods, where the lowest cost ($L.Cost$) and the average cost ($A.Cost$) over 5 runs for each problem are reported. In addition, the best-known lowest traveling cost of each CVRP instance is also shown via ``Optimal'' for reference. The ``$Gap$'' measures the difference between the gained result against the optimal solution, which is calculated via $(A-B)/B$, where $A$ and $B$ denote the obtained average traveling cost and the lowest traveling cost, respectively. Superior results gained by our proposed MLLM-V and the baseline MLLM-T have been highlighted using bold font.

\begin{table*}[htbp]
  \centering
  \caption{Numerical results gained by MLLM-T and MLLM-V where ``$A.Cost$'' and ``$L.Cost$'' represent the averaged traveling cost and the lowest traveling cost over 5 independent runs, respectively. Superior performance is highlighted in bold.}
  
  \setlength{\tabcolsep}{3.5mm}{
  \begin{tabular}{lrrrrrrr}
  \toprule
  \multicolumn{1}{l}{\multirow{2}{*}{Problem}} & \multicolumn{1}{c}{\multirow{2}{*}{Optimal}} & \multicolumn{3}{c}{MLLM-T} & \multicolumn{3}{c}{MLLM-V} \\
  &&$B.Cost$&$A.Cost$&$Gap$&$B.Cost$&$A.Cost$&$Gap$\\
  
  \midrule
P-n19-k2 & 213 & 263 & 292 & 31\% & 235 & \textbf{260} & \textbf{22}\% \\
A-n32-k5 & 788 & 1073 & 1161 & 47\% & 1040 & \textbf{1141} & \textbf{45}\% \\
A-n36-k5 & 802 & 1068 & \textbf{1210} & \textbf{51}\% & 1147 & 1229 & 53\% \\
A-n38-k5 & 734 & 981 & 1132 & 54\% & 982 & \textbf{1040} & \textbf{42}\% \\
A-n39-k5 & 829 & 1222 & 1366 & 65\% & 1112 & \textbf{1300} & \textbf{57}\% \\
A-n44-k6 & 939 & 1449 & 1568 & 67\% & 1319 & \textbf{1445} & \textbf{54}\% \\
A-n46-k7 & 918 & 1388 & 1566 & 70\% & 1430 & \textbf{1565} & \textbf{70}\% \\
A-n65-k9 & 1182 & 1979 & 2143 & 81\% & 1848 & \textbf{2124} & \textbf{80}\% \\
A-n69-k9 & 1166 & 2062 & 2161 & 85\% & 1807 & \textbf{2104} & \textbf{80}\% \\
E-n51-k5 & 525 & 897 & 1022 & 95\% & 825 & \textbf{881} & \textbf{68}\% \\
P-n55-k10 & 698 & 991 & 1045 & 50\% & 892 & \textbf{985} & \textbf{41}\% \\
P-n65-k10 & 797 & 1205 & 1269 & 59\% & 1098 & \textbf{1223} & \textbf{54}\% \\
P-n70-k10 & 830 & 1417 & 1558 & 88\% & 1218 & \textbf{1299} & \textbf{56}\% \\
X-n139-k10 & 13596 & 30763 & 42475 & 212\% & 28568 & \textbf{33163} & \textbf{144}\% \\
X-n143-k7 & 15697 & 46831 & \textbf{51238} & \textbf{226}\% & 38493 & 52339 & 233\% \\
X-n153-k22 & 21227 & 44070 & 53435 & 152\% & 39955 & \textbf{49732} & \textbf{134}\% \\
X-n162-k11 & 14139 & 33193 & 42320 & 199\% & 31518 & \textbf{38227} & \textbf{170}\% \\
  \bottomrule
  \end{tabular}
  }
  \label{tab:cvrp}
  \end{table*}
  
As can be observed from the table, MLLM-V achieves better results than MLLM-T on most of the CVRP instances. For instance, the average traveling costs attained by the proposed MLLM-V method on the ``P-n19-k2'' and ``A-n44-k6'' problems were 260 and 1445, respectively, exceeding the optimal solution of the MLLM-T method on these cases. This indicates that the utilization of visual prompts and their descriptions can significantly enhance the learning and optimization capabilities of the MLLM, resulting in more effective and efficient vehicle routing. Furthermore, by leveraging the rich multimodal information accessible from maps, traffic data, and other resources, MLLM-V can unveil new prospects for tackling challenging optimization problems with a handful of observations from similar solved problems, which is intractable to conventional machine learning methods.

\begin{figure}[htbp]
  \centering
  \subfigure[Optimal]{
  \includegraphics[width=0.45\columnwidth]{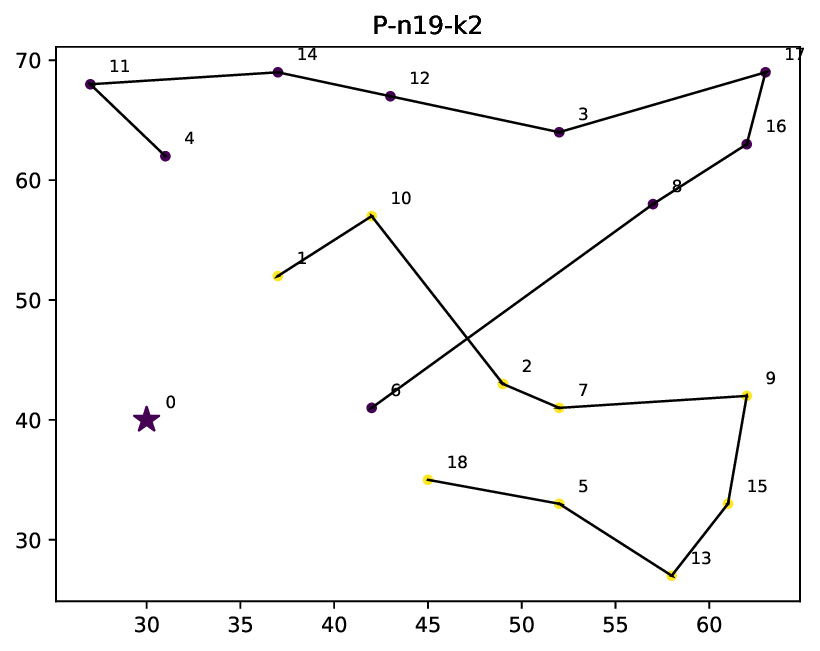}}
  \subfigure[Randomly Generated]{
  \includegraphics[width=0.45\columnwidth]{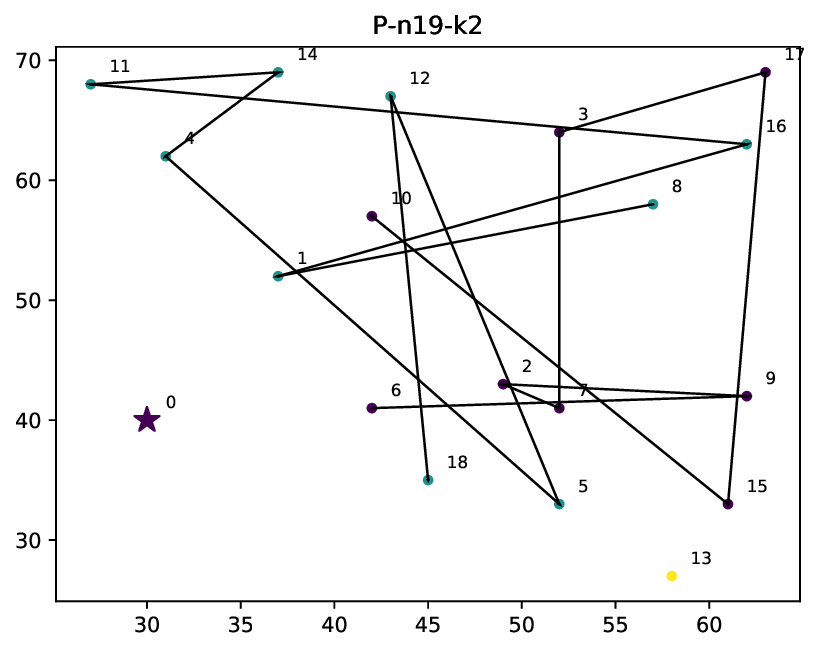}}
  \subfigure[Text Prompt Only]{
  \includegraphics[width=0.45\columnwidth]{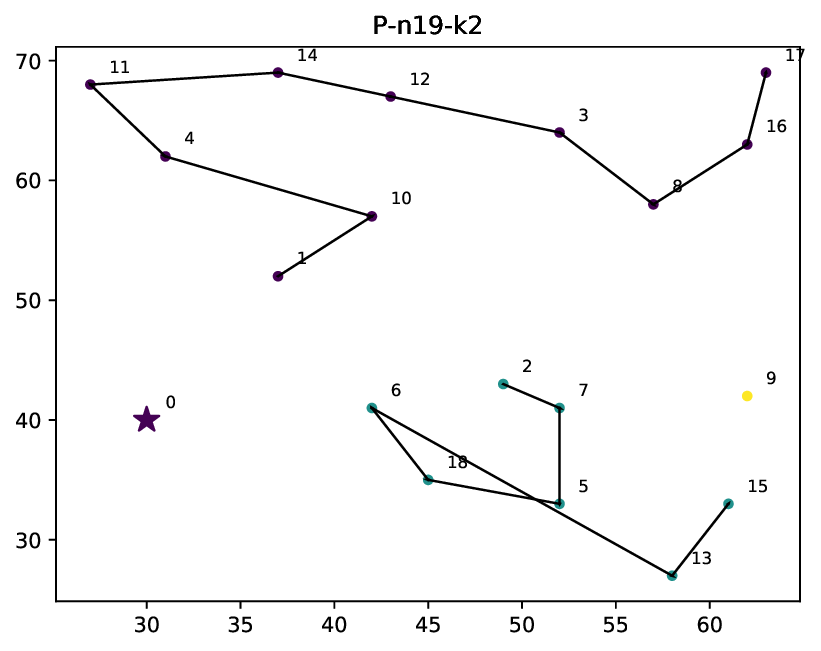}}
  \subfigure[With Vision Prompt]{
  \includegraphics[width=0.45\columnwidth]{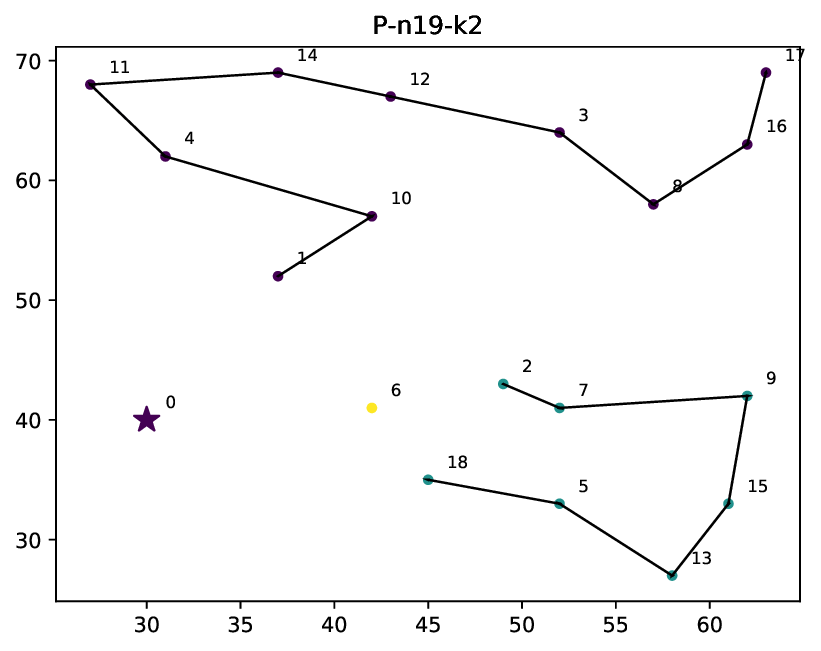}}
  \caption{Traveling routes obtained based on randomly generated, MLLM-T (i.e., ``Text Prompt Only''), and MLLM-V (i.e., ``With Vision Prompt''), respectively. ``Optimal'' shows the best routing solution of ``P-n19-k2''.}
  \label{fig:deep}
\end{figure}

To illustrate the differences between the optimization framework based on MLLM-V and MLLM-T, we compare the traveling routes for ``P-n19-k2'' generated by random method, MLLM-T with text prompt only and our proposed MLLM-T with both textual and visual prompts in Fig. \ref{fig:deep}. Furthermore, the optimal traveling routes of ``P-n19-k2'' are also provide for reference. As can be observed from Fig. \ref{fig:deep}, the optimal traveling routes (a) are concise and efficient paths connecting all points with minimal overlap and distance covered. In contrast, the randomly generated routes (b) exhibit a chaotic pattern with excessive overlaps, resulting in inefficient paths. The MLLM-T with text prompt only shows improvements over the random method but still has room for optimization as some overlaps occur. When we employ the MLLM-V that incorporates both textual and visual prompts (d) for vehicle routing, there is a noticeable enhancement in the optimization of traveling routes. As can be observed, the paths are more coherent, with reduced overlaps and structures more akin to the optimal routes. This reveals the efficacy of MLLM in enhancing optimization processes by leveraging both textual and visual data to generate more effective and efficient routing solutions.

Moreover, we have conducted a further investigation on the route construction processes for the ``P-n19-k2'' instance. The MLLM-T's response is given below:

\begin{center}
\fcolorbox{black}{cyan!20}{\parbox{0.9\linewidth}{
- \textbf{Route 1}: The route starts with customer 1 as it is one of the nearest customers to the depot. The vehicle then serves customers 10, 4, 11, 14, and 12 in sequence, following the principle of proximity and creating a path that leads away from the depot towards the top of the map. The vehicle then continues to customers 3, 8, and 16, which form a natural cluster in the top right corner. Finally, the vehicle visits customer 17, which is a bit further away but still in the same general direction, before heading back to the depot. The entire route forms a convex shape that encompasses the northern customers.\\
- \textbf{Route 2}: This route begins with customer 2, which is the closest to the depot among the remaining customers. The vehicle then visits customers 7, 5, and 18, which are all relatively close to each other and the depot, forming a small loop. Next, the vehicle heads to customer 6, which is a bit further but still on the way to customers 13 and 15, which are the farthest in this route. The vehicle serves customer 9 last before returning to the depot as it is the most easterly customer and forms a convex polygon when connected with the previous customers and the depot.\\
}}
\end{center}

The MLLM-V's routes construction process for the same instance is given below:
\begin{center}
\fcolorbox{black}{cyan!20}{\parbox{0.9\linewidth}{
-\textbf{Route 1}: The route begins with customer 1, who is nearest to the depot. From there, the vehicle serves customer 10, following the principle of proximity, and continues to customer 4, which forms a line segment with the previous customers. The route then extends to customers 11, 14, and 12, which are all located in the upper part of the map, close to each other. The vehicle then moves to the right, serving customer 3, and continues to 8, 16, and finally 17, which are all in close proximity, forming an almost convex shape. This route utilizes the vehicle's capacity well, with a total demand of 158, just below the capacity limit of 160.\\
-\textbf{Route 2}: This route starts with customer 2, who is relatively close to the depot but on the right side. The vehicle then serves customer 7, followed by 9, which are both nearby. The route extends to customer 15, who is further to the right, and then to 13, which is on the bottom right corner of the map. The vehicle then serves customer 5 and heads back towards the depot, stopping at customer 18 and finally at 6, which is close to the depot. This route also makes good use of the vehicle's capacity, with a total demand of 150, and covers customers in the lower and right-hand side of the map, ensuring that the vehicle's path is compact and efficient.
}}
\end{center}
As can be observed from the two responses, both methods (even without vertex distribution layout picture) can remember the accurate orientation of the vertexes in the first route. However, the MLLM-T failed to remember the orientation of the vertexes in the second route, thus resulting in a bad solution. In contrast, the MLLM-V can remember the accurate orientation of vertexes based on the distribution layout map, which demonstrates the advantages of using MLLM for optimization. Moreover, both of the methods failed to calculate the correct total demand of the second route, which again confirms that the MLLM is hard to remember accurate tail information as the demand is only given via text. 

\subsection{Why MLLM-V performs better than MLLM-T}
To give a deeper insight of the superiority of the proposed MLLM-V, we analyze the different heuristics captured by the model from both the textual and image observations. We compare them with the heuristics captured by the MLLM-T model, which only uses textual observations. We highlight the heuristics that are unique to the MLLM-V model in bold. The heuristics captured from the textual observations are given by:
\begin{center}
\fcolorbox{black}{cyan!20}{\parbox{0.9\linewidth}{
- Start by creating clusters of customers based on their proximity to minimize traveling distance within each route. \\
- Ensure that the total demand for each route is as close to the vehicle's capacity as possible without exceeding it. \\
- Consider the total traveling cost when constructing routes, aiming to keep it as low as possible. \\
- Allow for flexibility in route lengths to achieve overall efficiency. \\
- Make sure each customer is visited once and that all routes start and end at the depot. \\
}}
\end{center}

As can be observed, these heuristics are common to both the MLLM-V and MLLM-T models, as they are derived from the problem definition and the objective function, which reflect the basic principles of the vehicle routing problem, such as minimizing total traveling cost, while satisfying the demand and capacity constraints. On the other hand, the heuristics captured from both the textual and image observations are given by:
\begin{center}
\fcolorbox{black}{cyan!20}{\parbox{0.9\linewidth}{
- Starting with a clustering approach to group customers based on geographic proximity. \\
- Balancing the demand across routes to maximize vehicle capacity utilization. \\
- \textbf{Planning routes that begin and end close to the depot to minimize empty travel.} \\
- \textbf{Minimizing the overlap of routes to ensure distinct paths for each vehicle.} \\
- Considering the directness of travel between customers within the same route. \\
- \textbf{Strategically incorporating outliers into routes without significantly increasing costs.} \\
- Ensuring that edge cases are handled in a way that does not compromise the efficiency of other routes. \\
}}
\end{center}

The unique heuristics found by MLLM-V have been highlighted in bold font. As can be observed, by incorporating both textual and visual data, the proposed MLLM-V captures more advanced heuristics for vehicle routing than MLLM-T, which learn heuristics based on textual observations alone. These heuristics reveal the superiority of the MLLM-V model over the MLLM-T model, as they provide clear steps to construct cost-efficient traveling routes that lead to more effective outcomes.

\section{Conclusion}

In this paper, we presents a new approach to enhance the optimization performance of LLMs by incorporating multimodal inputs, i.e., text and images. We apply the MLLM to a classic combinatorial optimization problem (i.e., CVRP), and conduct empirical studies and analysis to demonstrate its effectiveness. Our experimental results reveal that MLLMs can grasp the inter-dependencies among decision variables and offer more profound insights into the optimization problem, as opposed to LLMs that depend solely on numerical text prompts. Our work illustrates the potential of MLLMs for tackling complex optimization problems and suggests promising directions for future research in this domain. 

% Despite the current challenges of multi-modal large model optimization in matching the heuristic algorithms devised by human experts, 

As future work, we would like to extend the application of MLLMs to other challenging optimization problems. Moreover, we aim to improve the optimization performance of MLLMs by generating heuristic functions from the observations, rather than directly producing solutions of problems encountered.

\bibliographystyle{named}
% \bibliography{resources/VRP,resources/LLM,resources/other,resources/traditional}
\bibliography{draft}

\end{document}